# Rethinking Fairness: An Interdisciplinary Survey of Critiques of Hegemonic ML Fairness Approaches

**Lindsay Weinberg**                                                    LWEINBER@PURDUE.EDU
*Honors College, Purdue University*
*West Lafayette, IN 47906*

## Abstract

This survey article assesses and compares existing critiques of current fairness-enhancing technical interventions in machine learning (ML) that draw from a range of non-computing disciplines, including philosophy, feminist studies, critical race and ethnic studies, legal studies, anthropology, and science and technology studies. It bridges epistemic divides in order to offer an interdisciplinary understanding of the possibilities and limits of hegemonic computational approaches to ML fairness for producing just outcomes for society's most marginalized. The article is organized according to nine major themes of critique wherein these different fields intersect: 1) how "fairness" in AI fairness research gets defined; 2) how problems for AI systems to address get formulated; 3) the impacts of abstraction on how AI tools function and its propensity to lead to technological solutionism; 4) how racial classification operates within AI fairness research; 5) the use of AI fairness measures to avoid regulation and engage in ethics washing; 6) an absence of participatory design and democratic deliberation in AI fairness considerations; 7) data collection practices that entrench "bias," are non-consensual, and lack transparency; 8) the predatory inclusion of marginalized groups into AI systems; and 9) a lack of engagement with AI's long-term social and ethical outcomes. Drawing from these critiques, the article concludes by imagining future ML fairness research directions that actively disrupt entrenched power dynamics and structural injustices in society.

## 1. Introduction

Recently, there has been a wave of AI scholarship working to define and measure fairness, equity, and bias in computational terms. While there is an array of computational approaches to fairness in the context of machine learning (ML), these approaches are typically through intervention prior to modeling, at the point of modeling, or after modeling, and involve metrics for quantifying the fairness of decisions and technical efforts for mitigating bias and unfairness (Caton & Haas, 2020). This article assesses and compares existing critiques of hegemonic fairness-enhancing interventions in ML that draw from a range of non-computing disciplines, including philosophy, feminist studies, critical race and ethnic studies, legal studies, anthropology, and science and technology studies (STS). It bridges epistemic divides in order to provide an interdisciplinary understanding of the possibilities and limits of these technical interventions for producing just outcomes for society's most marginalized. Drawing from a feminist STS perspective, the author understands AI as a sociotechnical system requiring an account of the ways that humans and AI tools are situated within social and historical contexts, as well as implicated in larger systems of power that shape the distribution of resources, opportunities, and (dis)advantages across the social field (e.g. militarism, capitalism, systemic racism). Power, in this sense, refers to the ability to coerce, oblige, influence, or command how life is organized and experienced by others through a range of methods. According to the insights of critical social theory, power is not simply an expression of individual actions and behaviors towards others, but a force that can be diffused through institutionalized ways of knowing, governing, and communicating about the world (Haugaard, 2002). Power is often distributed unevenly in society among individuals and groups through techniques of domination and subordination (e.g. surveillance, labor exploitation, racial discrimination). This results in different levels of access to resources and opportunities that social actors navigate within a given context, also known as power dynamics. As this article will argue, ML fairness is a discourse that is deeply embedded within larger institutional and social arrangements (structures) of power that actively shape our shared social world. Johnson (2005) explains discourse as "an institutionalized way of speaking that determines not only what we say and how we say it, but also what we do not say" (p. 1). To begin, then, it is worthwhile to start with how the debates concerning computational approaches to ML fairness are often framed.

ML fairness scholarship frequently incorporates a retelling of the 2016 ProPublica controversy concerning the COMPAS recidivism risk tool used in US courts (Mehrabi et al., 2021). This tool provides a prediction of the likelihood of a given defendant repeating a criminal offense compared to individuals from similar data groups. In this case, ProPublica researchers found that COMPAS disproportionately falsely labels Black criminal defendants in the US as high risk for committing future crimes. Northpointe, the owner of COMPAS, argued that the tool was fair because there was accuracy equity across racial groups. Ultimately, it became clear that due to unequal base rates between Black defendants and white defendants as a result of past patterns of historical discrimination, accuracy equity was mathematically impossible to achieve alongside equalized false positive rates (Green, 2018a). This debate, and the understanding of a range of irreconcilable "fair" statistical constraints, gave rise to benchmarks for assessing different statistical definitions of ML fairness, including equal opportunity, disparate mistreatment, statistical parity, counterfactual fairness, and disparate impact (Kleinberg et al., 2017; Green & Hu, 2018; Caton & Haas, 2020). Many of these benchmarks directly appropriate terms from US anti-discrimination law (Hoffman, 2019). These benchmarks have informed several open-source toolkits sponsored by major tech companies, including IBM's AI Fairness 360 for mitigating bias in training data (Varshney, 2018) and Google's ML-fairness-gym for anticipating the effects of automated decision systems (Srinivasan, 2020).

Many scholars have used the COMPAS debate to illustrate the limitations of hegemonic computational approaches to ML fairness. For instance, Green (2018a) demonstrates how the COMPAS tool itself is biased towards incapacitation because only incapacitation via recidivism has been rigorously measured in a manner conducive to ML, in contrast to many non-punitive alternatives. Furthermore, as Green and Hu (2018) argue, "even accurate and calibrated predictions of recidivism extend the legacy of historical discrimination by punishing blacks for having been subjected to…criminogenic circumstances in the first place" (p. 4). This fixation on parity or imparity between racial groups overlooks fairness considerations that cannot be captured statistically, such as whether the tool's fundamental purpose is itself fair (Green & Hu, 2018). The COMPAS debates have helped to entrench statistical notions of fairness into a range of ML fairness research efforts. Yet, these debates do not adequately capture the historical roots of contemporary institutional aspirations for algorithmic fairness.

In order to understand this phenomenon, Ochigame (2020) argues that we must first understand the emergence of the modern concept of risk. According to Ochigame (2020), throughout the nineteenth-century in the US, the concept of risk was intrinsically tied to the rise of corporate risk management, largely concerning "antebellum legal disputes over marine insurance liability for slave revolts in the Atlantic Ocean." Ochigame (2020) traces how the concept of risk was used to justify and entrench racial discrimination across a range of insurance industries throughout the nineteenth century. With the rise of modern mathematical statistics,

Fourcade and Healy (2013) argue that US postwar neoliberal era market institutions "increasingly use[d] actuarial techniques to split and sort individuals into *classification situations* that shape life-chances" (p. 559). By the second half of the twentieth century, actuarial methods were a key feature in policing, parole decisions, and credit bureaus. These methods were often couched in promises of transparency and equal, dispassionate treatment (Burrell & Fourcade, 2021). Furthermore, in the 1970s, credit-scoring companies used ideas of statistical objectivity in order to prevent anti-discrimination legislation from passing that would ban the use of these scores (Ochigame, 2020). When US civil rights and feminist activists challenged the use of risk classification in the pricing of insurance as being unfair and discriminatory, the insurance industry promoted the idea of actuarial fairness as a complex, technical, objective, and apolitical matter in order to evade regulation (Ochigame, 2020). This move effectively reframed the debate about the fairness of insurance around inaccurate predictions, rather than the use of actuarial risk classification itself.

Furthermore, scholars like Hutchinson and Mitchell (2019) and Hoffman (2019) have critically evaluated the influence of 1960s US antidiscrimination discourse on contemporary AI fairness debates. This discourse has significantly shaped how ML fairness is understood legally, institutionally, and technologically. Hutchinson and Mitchell (2019) describe how 1960s and 1970s debates from the use of assessment tests in public and private industry "stimulated a wealth of research into how to mathematically measure unfair bias and discrimination within the educational and employment testing communities, often with a focus on race" (p. 1). These research efforts included the "introduction of formal notions of fairness based on population subgroups, the realization that some fairness criteria are incompatible with one another, and pushback on quantitative definitions of fairness due to their limitations" (p. 1). In the 1980s, there were significant fairness debates concerning educational testing, and particularly, whether educational testing reproduced and intensified racial inequality, or whether group-based fairness considerations unfairly privileged members of minority groups (Hutchinson and Mitchell, 2019). According to Hutchinson and Mitchell (2019), over time, fairness determinations became "less tied to the practical needs of society, politics and law, and more tied to unambiguously identifying fairness" (p. 9). These efforts to unambiguously identify fairness served to depoliticize fairness metrics and concealed the ways that researchers' values, goals, and assumptions about the world shaped fairness assessments.

Based on AI fairness debates' relationship to anti-discrimination discourse, Hoffman (2019) argues that current research on AI fairness often inherits this discourse's limitations for realizing social justice. As she explains, "in mirroring some of antidiscrimination discourse's most problematic tendencies, efforts to achieve fairness and combat algorithmic discrimination fail to address the very hierarchical logic that produces advantaged and disadvantaged subjects in the first place" (p. 901). Hoffman (2019) revisits critiques of US antidiscrimination law from the 1970s, including the ways that antidiscrimination discourse shifted away from broader aims of social justice through an emphasis on: discrete "bad actors" and individual prejudices rather than structural causes; single-axis approaches to discrimination rather than looking at how forces of gender-, race-, ability-, and class-based oppressions are produced and intersect; a focus on relative disadvantage rather than the systematic production of advantages for privileged groups; and a focus on a limited set of fairness considerations, such as the distribution of a narrow set of goods rather than the conditions that structure the distribution and exercise of rights, opportunities, and resources in the first place.

By widening the frame of ML fairness debates to include these earlier periods, we can understand contemporary debates as the latest iteration of longstanding struggles for just sociotechnical arrangements. Furthermore, this frame demands an engagement with the ways that ML fairness discourse itself is actively produced within existing social, political, and historical realities, the power dynamics of which shape hegemonic approaches to ML fairness in the present, and often remove from view relevant structural and social concerns for a more just society. Thus, the central goal of this survey article is to summarize and assess critiques of contemporary, hegemonic ML fairness research in order to help AI researchers broaden approaches for disrupting these entrenched power dynamics, including who generally gets to participate in the design of ML tools, how, and who benefits from their deployment. In the following section, the author describes how the relevant literature was identified. Section 3 gives the author's breakdown of the major themes of critique, along with additional details, explanation, and examples from the literature. Section 4 provides an overview of proposals for how the ML fairness community can address these critiques, and evaluates whether these proposed solutions are likely to disrupt entrenched power dynamics and structural injustices in society. The article concludes in Section 5 with a discussion of what an anti-oppressive, power-centered approach to ML fairness research could prioritize in order to foreground social justice.

## 2. Article Selection

In order to conduct this survey, the author limited the search criteria to papers, articles, books, and conference proceedings published after 2015 with full text available, and that explicitly position themselves as critiques of computational fairness interventions in ML.[1] While issues of fairness in sociotechnical systems were considered long before 2015 across a range of disciplines, including philosophy, computer science, law, and STS, this article sets out to map the contemporary landscape of critiques of fair-ML in particular, which expanded substantially in the wake of the COMPAS debates. For conference proceedings, the author focused primarily on the Association for Computing Machinery (ACM) Conference on Fairness, Accountability, and Transparency, given its multidisciplinary scope and fairness focus. After identifying relevant literature available through the author's institutional library, Google scholar, and arXiv through a title and abstract screening process, the author then used Zotero in order to manually tag and annotate relevant readings according to the key thematic concerns (e.g., technological solutionism) epistemic frameworks (e.g., feminist standpoint theory), and core theoretical concepts (e.g., data justice) that emerged in the full text. This database ultimately included a total of 58 sources published between 2015 and 2021. The author found that there were nine major themes of critique running through the sampled scholarship concerning the following: 1) how fairness gets defined; 2) how problems for AI systems to address get formulated; 3) the impacts of abstraction on how AI tools function and its propensity to lead to technological solutionism; 4) how racial classification operates within AI fairness research; 5) the use of AI fairness measures to avoid regulation and engage in ethics washing; 6) an absence of participatory design and democratic deliberation in AI fairness considerations; 7) data collection practices that entrench "bias," are non-consensual, and lack transparency; 8) the predatory inclusion of marginalized groups into AI systems; and 9) a lack of engagement with AI's long-term social and ethical outcomes.

It is worth noting the epistemological and evaluative assumptions embedded in the NSF Program on Fairness in Artificial Intelligence in Collaboration with Amazon, which funded this survey research as part of a larger cross-institutional AI fairness project. The grant solicitation's framing asserts that past NSF-funded innovations in AI and ML have offered "new levels of economic opportunity and growth, safety and security, and health and wellness, intended to be shared across all segments of society." Furthermore, the grant solicitation argues that there is a need to encourage broad acceptance and adoption of large-scale deployments of AI so that these benefits can be fully enjoyed, and that this can be accomplished through the creation of "trustworthy" AI tools that are fair, transparent, explainable, and accountable. Thus, the explicit goal of this NSF program is to fund computational research that will promote the acceptance of AI systems, and AI's capacities for social good are assumed a priori. The author sees this assumption as a feature of the ways that power dynamics shape knowledge claims, the research that gets funded and given access, and finally, who generally gets to conduct this research. This survey article operates in productive tension with these assumptions and goals in order to map modes of AI fairness research inquiry that are not skewed towards the priorities of individuals and institutions that benefit from structural injustice, and to expand dialogue between AI/ML fairness practitioners and critical scholars operating within and outside of computing fields.

---

[1] In some cases in the surveyed scholarship, authors extend their critiques of ML to the broader AI community and/or use AI as an umbrella term including ML. This is reflected in the article with "AI/ML" and "AI" respectively when needed.

## 3. Major Themes of Critique

### 3.1 How Fairness Gets Defined

According to Green and Hu (2018), the hegemonic theory of fairness in ML is that fairness can be presented in terms of a domain-general procedural or statistical guideline, and that "this definition can be operationalized such that, so long as the chosen fairness criteria are satisfied, the resulting procedures and outcomes of the system are necessarily fair" (p. 1). In contrast, the authors argue that fairness judgments are always under contestation, and that all fairness measures in ML involve "value-laden assertions about a system's purpose, individuals' entitlements, and the relevant criteria for decision-making" (p. 2). For Green and Hu (2018), equating statistical parities of various types to fairness limits the realm of fairness considerations and side-steps questions such as, is the tool's fundamental purpose a just one? Are individuals' entitlements to fair procedure respected? Similarly, the authors are concerned with the ways that quantifiable metrics can get prioritized over other value considerations, as well as a lack of engagement with possible downstream effects and the need for democratic deliberation (see section 3.7). To give an example, the authors describe how COMPAS's prioritization of recidivism rates could have the unintended consequence of leading judges to place greater emphasis on incapacitation as the goal of sentencing, which then leads to significant policy and jurisprudence shifts.

This concern with the reduction of fairness considerations to quantifiable metrics is shared by scholars like Ochigame et al. (2018), who argue that "any claim of fairness requires a critical evaluation of the ethical and political implications of deploying a model in a specific social context" (p. 1). In order to illustrate this argument, Ochigame et al. (2018) use the example of risk assessment algorithms in the US penal system, highlighting a range of concerning features. According to these authors, AI fairness researchers engaging with these tools often claim that models can predict the likelihood that an individual will commit a crime when the data does not represent the prevalence of crimes, but rather, recorded arrests, convictions, or incarcerations, which leads to inaccuracies and fallacious claims about crime prediction. Additionally, researchers conceptualize interpretability in terms of listing a sequence of steps taken, rather than documenting background assumptions and domain specific theories that can be deeply problematic, such as flawed criminological theories that are biased towards mass incarceration, and the use of static factors like age, sex, and arrest history as ostensible measures of "risk." Furthermore, researchers turn fairness into a mathematical property of algorithms, which is then used to designate a given "solution" as fair, precluding approaches to penal reforms that focus on decarceration. These reforms include the elimination of monetary bail, the abolition of mandatory minimum sentences, and the reduction of prison admission and sentence lengths. Ultimately, Ochigame et al. (2018) argue that fairness should be understood not as a property of algorithms but rather of decisions, "justified by interpretations of data analyses, in particular social contexts, from the perspective of ethical and political commitments of specific subjects" (p. 4). Furthermore, they encourage vigilance regarding the ways that ideas of fairness are used to promote "the false and dangerous claim that fairness is a technical and apolitical matter," (p. 5) resulting in the legitimization of the status quo.

Binns (2018) argues that fairness is used as a placeholder for a range of egalitarian considerations in ML. Regarding discrimination, the author addresses understandings of discrimination as either a decision-maker's bad intent or animosity, or as the existence of disparate impact, wherein decision-makers fail to anticipate or redress disparities in people's treatment. He argues that mental state accounts of discrimination would be difficult to apply to the context of algorithmic decision-making, given that AI tools do not bear states such as contempt or animosity, although data can reflect either the intentional encoding of bias or negligence in addressing bias. Ultimately, Binns (2018) highlights the "practical challenges which may limit how effective and systematic fair ML approaches can be in practice" (p. 9), given that it is difficult to assess an individual's responsibility, culpability, or desert without extensive information, "such as their socio-economic circumstances, life experience, personal development, and the relationships between them" (p. 9). This paper points to the limitations of training data and legally protected categories for doing justice to complex and highly contingent and contested fairness considerations.

Furthermore, Mitchell et al. (2021) argue that mathematical definitions of fairness treat the learning problem that is used to formulate a predictive model as external to the evaluation of fairness, which can overlook issues concerning sampling bias, societal bias, and temporal dynamics. The authors also identify utilitarian assumptions generally built into predictive models, including the assumptions that outcomes are not affected by the decisions of others, that social welfare is a sum of individual utilities, and that individuals can be considered symmetrically, e.g., the harm of denying a loan to someone who could repay is equal across people. The authors contend that, at best, formal fairness metrics can be used to illustrate "when interventions are necessary to bring about a different world" (p. 158), and to recognize trade-offs that don't need to be accepted as a given, but can rather be used to encourage the consideration of a range of options beyond predictions to realize ethical goals.

## 3.2 Problem Formulation

A second line of critique of concerns how problems for ML to address get formulated in the first place. Passi and Barocas (2019) argue that greater attention needs to be placed on the practice of problem formulation in data science in order to better understand the ethical concerns that emerge at this stage, including: a potential bias towards what is most easily quantifiable; target variables that produce unfairness because they vary by population; and problem formulations that are based on objectionable ways of achieving desired outcomes. To illustrate this last point, they offer the example of financial aid admissions strategies that maximize the recruitment of wealthy and/or "gifted" students using multiple smaller offers, rather than using those resources to support marginalized applicants. Different potential proxies frame a given problem in subtly different ways and raise different ethical concerns. Thus, problem formulation necessitates a careful discussion about normative commitments, meaning one's values, assumptions, and priorities concerning what ought to be.

Similarly, Mitchell et al. (2021) argue that the social objective of deploying a model, the individuals subject to the decision, and the decision space are all value-laden choices that impact fairness. They point to the ways that certain design choices can lead to poor outcomes, such as narrowing focus to a chosen, measured outcome (omitted payoff bias). They also describe the risks of only evaluating fairness criteria on the population to which the model is applied, which may overlook unfairness in terms of who came to be subject to the model in the first place.

Furthermore, restricting interventions/options within the decision space can have negative social and ethical implications. One such example is limiting the pre-trial decision space to either release, set bail, or detain, as opposed to directing a person to expanded pretrial services, or focusing on individual as opposed to community-based policies more broadly.

Fazelpour and Lipton (2020) argue that the key for determining what should be optimized for in a given context is to base one's understanding on the relevant causal mechanisms that account for present injustices and that govern the impact of proposed interventions. These authors push for evaluating the fairness of ML using a sociotechnical framework that accounts for social context, actors, users, and dynamic impact, which necessarily goes beyond "simple modifications of existing ML algorithms" (p. 11). According to these authors, who draw from political philosophy, ideal theories—theories that begin with a conception of an ideally just world under ideal conditions, which then provide a target state and evaluative standard—seem to implicitly inform much of the current work on fairness in ML, and are in need of challenging for the following reasons. Ideal theories "(1) can lead to systematic neglects of some injustices and distort our understanding of other injustices; (2) do not, by themselves, offer sufficient practical guidance about what should be done, sometimes leading to misguided mitigation strategies; and finally (3) do not, by themselves, make clear who, among decision-makers is responsible for intervening to right specific injustices" (Fazelpour & Lipton, 2020, p. 3). In contrast, the non-ideal approach begins by identifying actual injustices that will likely impact those on the receiving end of a decision-maker's behaviors, which requires evaluating what causes the problem and the responsibility of different actors to address it. These authors argue that many of ML researchers' primary approaches fail to consider broader social contexts, rely on reductive decontextualized parity-constrained formulations of fairness, and neglect considerations about the responsibilities of the decision-maker as well as the impact of a proposed intervention within the existing social system.

Overdorf et al. (2018) also point to the ways that the goals of optimization systems are in need of greater scrutiny within discussions of ML fairness. They argue that calls for fairness often overlook the fact that these goals might intentionally "aspire for asocial or negative outcomes," and "lack incentives to maximize society's welfare" (p. 3). Thus, providers might simply utilize fairness solutions that align with their goals. For example, Overdorf et al. (2018) describe how Uber could apply fairness metrics to their rider-driver matching algorithm with the goal of ensuring that all Uber drivers are equally likely to find riders, access surges, and earn similar wages across protected categories, without addressing the ways their algorithm suppresses driver wages in order to maximize profit. Furthermore, these authors are concerned with the failures of service providers to honestly and effectively apply fairness solutions. There are frequent conflicts between fairness and profit maximization, and service providers are often focused on outcomes rather than processes. This results in service providers ignoring or overlooking the substantial and often negative "externalities" they create. These negative features include: disregard for non-users and environments; disregard for non "high-value" users; externalization of exploration risks to users and environments; unfair distribution of errors; prioritizing optimization goals through shortcuts; and concentration of resources, power, and data collection practice in a few data holders. Similarly, Kasy & Abebe (2020) emphasize, "leading notions of fairness take the objective of the algorithm's owner or designer as a normative goal" (p. 1) and ignore within-group inequalities. This leads to the perpetuation of inequalities that are justified on the grounds of "merit," which reinforces the status-quo distribution of power between decision-makers and those who are impacted. Furthermore, like Overdorf et al. (2018), they note that limiting fairness considerations in the narrow context of the algorithm's immediate application overlooks its impact on the wider population.

### 3.3   Abstraction and Technological Solutionism

A third line of critique regarding hegemonic computational approaches to ML fairness is abstraction, meaning the ways that ML fairness gets conceptualized such that AI tools are removed from the social contexts in which they are generated and applied, ultimately resulting in the loss of necessary information for understanding and implementing fairness (Selbst et al., 2019). Selbst et al. (2019) understand fairness not as a property of technical tools, but of social systems, and identify five failures that result from computer scientists not considering social context: the framing trap, meaning the "failure to model the entire system over which a social criterion, such as fairness, will be enforced," (p. 60); the portability trap, meaning the "failure to understand how repurposing algorithmic solutions designed for one social context may be misleading, inaccurate, or otherwise do harm when applied to a different context" (p. 61); the formalism trap, meaning the "failure to account for the full meaning of social concepts such as fairness, which can be procedural, contextual, and contestable, and cannot be resolved through mathematical formalisms" (p. 61); the ripple effect trap, meaning the "failure to understand how the insertion of technology into an existing system changes the behaviors and embedded values of the pre-existing system" (p. 62); and finally, the solutionism trap, meaning the "failure to recognize the possibility that the best solution to a problem may not involve technology" (p. 63).

   Abstraction is also a key issue for Green and Viljoen (2020), who argue that abstraction can be problematic in terms of a loss of engagement with the complexity of the existing social and political conditions that shape AI tools, designs, and outcomes. For Green and Viljoen (2020), formal approaches to algorithmic fairness "fail to provide the epistemic and methodological tools necessary to fully identify and act upon the social implications of algorithmic work" (p. 20). They substantiate this claim using the example of hegemonic approaches to algorithmic fairness for predictive analytics in the US criminal legal system. They point to this work's limited framing of tradeoffs, a lack of critical engagement with what counts as a "crime" for predictive policing algorithms, tendencies towards technological solutionism and determinism, as well as a lack of explicitly articulated normative commitments. According to these researchers, abstraction contributes to the reasons why formal approaches to algorithmic fairness "typically overlook the ways in which these 'fair' algorithms can lead to unfair social impacts, whether through biased uses by practitioners, entrenching unjust policies, distorting deliberative processes, or shifting control of governance towards unaccountable private actors" (Green & Viljoen, 2020, p. 24). Furthermore, Green and Viljoen (2020) argue that the sense that algorithms can be applied to all situations and problems means that its methods often dominate and crowd out other forms of knowledge and inquiry, and particularly local forms of situated knowledge that may be better positioned to address a given task. Situated knowledge, drawing from feminist epistemology, thinks of knowledge as a contextual practice, wherein the social locations of individuals shape how they come to "know" the world (Haraway, 1988).

   It is important to note that abstraction is deeply embedded in dominant epistemological and educational frameworks in computer science (CS). Malazita and Resetar (2019) argue that the epistemological practices of the computer science discipline and pedagogy privilege instrumental and abstracted programming techniques. Technical competency is often "split from the social, political, and ideological world" (p. 301). They argue that this split is doubly impactful for marginalized students, who then have to embody knowledge practices and epistemological frameworks that counter their lived experiences. These students are actively recruited into the very CS practices that disparately harm marginalized groups. Furthermore, these epistemic tendencies provide narrow opportunities for students to articulate the negative consequences of these tools in the classroom. According to the authors, abstraction in CS intentionally limits knowledge of the broader system coding practices are situated within for the purposes of efficiency and achieving specified goals, which problematically constrains what counts as relevant points of intervention.

## 3.4   Racial Classification in AI Fairness

The challenges of abstraction are intertwined with the limitations of existing hegemonic approaches to racial classification in ML fairness. As Benthall and Haynes (2019) argue, abstraction contributes to how racial categories for fairness considerations get treated as "nominal categories of personal identity" rather than as qualities that are part of "systems of hierarchical social statuses" (p. 1). In other words, they argue that race is best understood as a political category rather than as "an intrinsic property of persons" (p. 4). These researchers offer a brief history of how racial categories have been constructed and contested to show their inextricable connection to systems of racial inequality. Racial categorization is inseparable from the systems of segregation and stratification in housing, education, employment, and civic life that allow for racially marked individuals to be sorted and then "subject to disparate opportunities and outcomes" (p. 2).

Hanna et al. (2020) also argue that algorithmic fairness research does not adequately account for the ways that social group categories, which AI fairness research operationalizes, are socially constructed, resulting in the "widespread use of racial categories as if they represent natural and objective differences between groups" (p. 2). They demonstrate how mathematically comparable group-based fairness criteria, such as parity-based fairness metrics, often treat oppressed social groups as interchangeable and deny the "hierarchical nature and the social, economic and political complexity of the social groups under consideration" (p. 8). They draw from the work of Black feminist scholarship and critical race theory's critique of fairness within liberal democracies to highlight how simplistic and decontextualized understandings of race obscure the unique oppressions different groups encounter. As fields of knowledge, the Black feminist tradition highlights the interlocking oppressions of racism, sexism, and economic subordination that condition Black women's lives (Collins, 2000), and critical race theory interrogates how these interlocking oppressions are embedded in purportedly color-blind laws and institutions (Crenshaw, 2019). Ultimately, this accepted approach to racial classification within hegemonic ML fairness research minimizes the structural factors that contribute to algorithmic unfairness.

## 3.5   Regulation Avoidance and Ethics Washing

Another key theme concerning critiques of hegemonic algorithmic fairness metrics is that these measures are intertwined with big tech's interests in preventing the emergence of regulations that significantly curtail or ban the deployment of AI/ML for commercial and military purposes. Drawing from his time as a graduate student researcher on ethical AI at the MIT Media Lab, Ochigame (2019) offers examples of how academic AI ethics work is used to influence policy in ways that ethics wash a range of AI applications, including the use of algorithms for detention decisions and drone warfare. Furthermore, Ochigame (2019) argues that academic AI ethics research has helped shift the discussion towards industry self regulation, including "voluntary 'ethical principles,' 'responsible practices,' and technical adjustments or 'safeguards' framed in terms of 'bias' and 'fairness' (e.g., requiring or encouraging police to adopt 'unbiased' or 'fair' facial recognition)." He also describes how some larger firms prefer mild regulation to none at all, as they have the means to more easily comply with regulatory requirements when compared to competitors, and can therefore consolidate market control. Ultimately, Ochigame (2019) posits that, "no defensible claim to 'ethics' can sidestep the urgency of legally enforceable restrictions to the deployment of technologies of mass surveillance and systemic violence." While some scholars like Barocas (2019) do not go as far as to say that AI ethics discourse is co-produced with corporate efforts to avoid regulation, Barocas (2019) does feel that companies have co-opted the fair ML community's efforts in order to justify greater data collection and oversimplify fairness issues in ways that align with their interests, rather than with progressive goals.

## 3.6 Absence of Participatory Design and Democratic Deliberation

These literatures' efforts to shift hegemonic ML fairness discourse towards more progressive goals often include a critique of the absence of participatory design and democratic deliberation in algorithmic fairness work. Participatory design refers to a design approach that involves users in the design process for the purposes of ensuring that a given tool is informed by, and responsive to, the needs of the people/groups who will be impacted by its use (Irani et al., 2010). Participatory design and democratic deliberation are valuable to ML fairness research, given that many ML tools have disparate impacts on marginalized groups, and given that fairness judgments are always under contestation. Furthermore, Green and Hu (2018) explain how the absence of democratic deliberation allows for a given ML tool to have downstream effects (see section 3.9) that then lead to significant political and institutional shifts without meaningful debate or legal opportunities for contestation.

According to Blodgett et al. (2020), the absence of participatory design is one of the ways that dominant power relations between technologies and racialized communities get reproduced. As the authors explain, "algorithmic fairness techniques, by proposing incremental technical mitigations—e.g., collecting new datasets or training better models—maintain these power relations by (a) assuming that automated systems should continue to exist, rather than asking if they should be built at all, and (b) keeping development and deployment in the hands of technologists" (Blodgett et al., 2020, p. 7). The absence of members from groups who are disadvantaged or multiply-burdened by systems of power in design processes often contributes to heightened levels of scrutiny and increased exposure to potential harms for marginalized people (Costanza-Chock, 2018). Furthermore, according to Kalluri (2020), the failure to "listen to, amplify, cite, and collaborate with communities that have borne the brunt of surveillance" has contributed to the presupposition, dominant within computational approaches to ML fairness, that AI is "marred only by biased data." While Cowgill et al. (2020) set out to test whether biased predictions arise from biased programmers or biased datasets, their heuristic is whether AI engineers from different demographic backgrounds will make technical adjustments for fairness in their code when designing ML tools for hiring decisions (see section 3.7 for limits of technical debiasing methods). This approach, as Barocas et al. (2019) note, overlooks a range of critical questions that marginalized people might bring to bear in a given AI application context. Instead, Kalluri (2020) argues that AI fairness practitioners should be asking how the field is currently conditioned and constrained by dominant power relations that actively shape research agendas.

In terms of the absence of participatory design, Malik (2020) is careful to distinguish perfunctory non-binding consultation from the meaningful participation of impacted groups in the design of ML tools, which includes their participation in the problem formulation, design, labeling, refinement, and testing stages with veto power. There are numerous studies where the inclusion of socially and economically marginalized groups is done through the use of services like Amazon's Mechanical Turk, wherein input from these groups is collected about the fairness of a given AI outcome, but where those human subjects are otherwise not engaged in the AI tool's development, nor empowered to effect any changes. Robertson and Salehi (2020) demonstrate how the elicitation of preferences over a fixed set of alternatives is often insufficient and unequal, given the ways individual preferences typically reflect existing social biases and inequalities, and given that these methods are not always paired with meaningful democratic deliberation. Additionally, Whittaker et al. (2019) describe how, in the context of AI for disabled people, these community members are frequently excluded from the design process itself, and often lack the ability to access, modify, and hold to account the technology marketed to them. It is also noteworthy that in Madaio et al.'s (2020) semi-structured interviews and workshops for co-designing AI fairness checklists with 48 ML practitioners, they found that participants reported that soliciting input and concerns from diverse stakeholders on AI fairness, including getting feedback from groups with significantly less power or influence than others, was an area that they felt would be difficult to integrate into their existing workflows due to a lack of resources and gaps in existing UX research methods. Furthermore, in a global context and drawing from Jobin et al. (2019), Mohamed et al. (2020) document how geographic areas like Africa, South and Central America, and Central Asia and their structural legacies of colonialism are underrepresented in AI ethics debates. These legacies include exploitative medical and scientific experimentation on colonized peoples, the appropriation of land and labor, and control over social structures in order to oppress colonized populations (Mohamed et al. 2020, see also Rodney, 2018). The authors argue that AI developmentalist projects are often rooted in paternalism, technosolutionism, and predatory inclusion (see section 3.8) rather than in intercultural dialogue and solidarity with grassroots organizations and marginalized people. This leads to practices where overdeveloped countries export harms and unethical AI research practices to vulnerable populations in low- and middle-income countries with weaker protections, perpetuating global power dynamics of experimentation and extraction rooted in colonialism (Mohamed et al. 2020).

## 3.7   Data Collection and Bias

Some scholars have also called for greater attention to the ethics of data collection underpinning ML fairness research, including the ways that cultural values and social and historical contingencies get embedded into data collection, cleaning, curating, and management (Elish & boyd, 2017). Others have critiqued the ways that improving the accuracy and thus "fairness" of algorithmic systems often involves surveillance, without people's consent, as well as the exploited labor of people who manually classify objects for training purposes (Kulluri, 2020; Sloane et al., 2020). Furthermore, Mohamed et al. (2020) call attention to how colonial power relations—the reproduction of structural inequalities that can be "contextualized historically as colonial continuities" (p. 8)—manifest in AI design and deployment, including through the conditions of "ghost workers" who annotate data. In many cases, this work is done by prisoners and the economically vulnerable, and in geographic areas with limited worker protections.

Additionally, a range of scholars have sought to problematize the focus on data "bias" as a means of addressing algorithmic unfairness. Powles and Nissenbaum (2018) argue that the problem of AI bias, and the fixation on notions of measurable, mathematical ideas of fairness to address it, has "limited the entire imagination of ethics, law, and the media as well." They identify three problems with a focus on AI bias. The first problem is that understanding bias as a computational problem rather than a social problem obscures its root causes. The second problem is that addressing bias can have perverse consequences, which Powles and Nissenbaum (2018) illustrate using the example of improving the accuracy of facial recognition systems for women of color. This approach to addressing bias can increase this group's susceptibility to surveillance and classification (see section 3.8). Thirdly, a focus on bias distracts from other important and larger considerations, such as the "colossal asymmetry between societal cost and private gain in the rollout of automated systems" (Powels & Nissenbaum, 2018). This focus on AI bias frequently frames AI as inevitable, rather than a tool that often emerges from power asymmetries in society. Dobbe et al. (2018) also urge for a shift away from a fixation on data bias, and towards the ways that the very methods and approaches that the ML community uses to reduce, formalize, and gather feedback are themselves sources of bias. Amoore (2020) also takes up this point about epistemology, arguing that ML algorithms "categorically require bias and assumption to function in the world" (p. 74).

Similarly, West, Whittaker, and Crawford (2019) argue that AI fairness research needs to go beyond technical debiasing to include a wider social analysis of how AI is used in context so that a more capacious accounting of bias is possible. They explain, "harms are frequently defined in terms of economic disadvantage, with less attention paid to harms of how people are represented or interpreted by AI systems, and the downstream social and political consequences of such representation" (p. 16). Furthermore, they argue that attempts to address AI unfairness by locating individual biases within a given technological system, and attempting to fix these biases by tweaking that system, overlooks the context of existing power dynamics. These power dynamics shape how a given AI tool will impact the distribution of benefits and harms in society. Ultimately, while a given AI tool's application may be free from bias in a statistical sense, this does not mean it is fair in an ethical sense (Mitchell et al., 2021). Drawing from the Black feminist tradition, Hampton (2021) argues that "oppression" rather than "bias" is a better descriptor for AI tools that contribute to material harms against marginalized people. "Oppression" helps link ML fairness discourse to the longstanding history of scientific and technological abuses that systematically advantage dominant groups. These groups continue to hold disproportionate power over how AI is being developed, assessed, and debated (see section 4.12).

## 3.8  Predatory Inclusion

The issue of AI bias is intrinsically tied to the issue of the predatory inclusion of marginalized groups into AI systems, given the ways that attempts to debias AI can be predicated on unethical data collection practices and/or expose marginalized groups to increased surveillance (West et al., 2019; Bennett & Keyes, 2019). For instance, videos from transgender YouTubers were taken without their knowledge for the purposes of training facial recognition technology (FRT) to better recognize people in the process of transitioning (West et al., 2019). Another example is the improvement of the "fairness" of FRT for people of color. Existing forms of FRT have been shown time and again to have higher error rates for women, people of color (Bualamwini & Gebru, 2018), and transgender people (Keyes, 2018). However, the inclusion of marginalized people into facial recognition datasets ultimately justifies the widening of surveillance's net, and by proxy, the reach of systems of oppression that deploy forms of automated recognition. Browne's (2015) work has demonstrated the deeply entrenched historical linkages between surveillance and racial oppression, from the transatlantic slave trade to contemporary forms of racial profiling in Canada and the US. More accurate facial recognition will not rectify criminal legal systems that are historically rooted in racial oppression. For instance, US laws and policing practices concerning what is treated as "criminal" behavior, and how, have continuously resulted in the oppression of communities of color, from slave patrols, to the enforcement of Jim Crow-era systems of racial segregation, to the War on Drugs (Davis, 2003; Gilmore, 2007; Benjamin, 2019). Furthermore, images of students, immigrants, abused children, people who have had mugshots taken, and deceased people have all been used without consent to test facial recognition technology within and outside of the US (e.g., Keyes et al., 2019; Harvey & LaPlace, 2019). Sloane et al. (2020) have also cautioned against "participation-washing," wherein powerful agencies use the participation of communities to manufacture consent and legitimize injustice. As these authors note, this is a longstanding practice in the context of the international development sector that has particular relevance for ML. Some approaches to participatory design (see section 4.9) can lead to predatory inclusion, wherein the labor and data from marginalized groups are extracted for making oppressive AI tools "fairer."

### 3.9 Lack of Engagement with Long-Term Outcomes

The final thread of critique running through these literatures concerns the ML fairness community's lack of engagement with the long-term outcomes of AI systems. As Green (2019) explains, there is a prioritization of immediate improvements over long-term impacts. This limits the degree to which unintended impacts, as well as alternative solutions that "might be necessary in conjunction with or instead of algorithms" (p. 4), are considered. Similarly, Green and Hu (2018) identify a lack of engagement with possible downstream effects and thus unintended consequences that can result from the implementation of a given ML tool, such as COMPAS's potential to lead judges to "place greater emphasis on incapacitation as the goal of sentencing" (p. 3). However, as Parvin and Pollock (2020) caution, the term "unintended consequences" in science and technology discourses is often used to dismiss important ethical and political considerations. This term sidesteps the fact that, in many cases, the consequences could have indeed been anticipated in advance by those from specializations that are often not conceived of, or empowered to, participate in the design and implementation of technology. Additionally, Parvin and Pollock (2020) argue that "phenomena described as unintended consequences are deemed too difficult, too out of scope, too out of reach, or too messy to have been dealt with at any point in time before they created problems for someone else" (p. 322). Thus, the term can function as a defensive and dismissive strategy that allows for the abdication of developer responsibility and accountability, on the one hand, and a lack of "consideration of the complexity of social systems in such a way as to lead to quick technical fixes" (p. 326), on the other. Jasanoff (2016) also argues that the term "unintended consequences" functions as a means for technologists to abdicate responsibility and minimizes the need for imagining negative implications in advance.

Additionally, Elish and boyd (2017) have pointed to the ways that celebratory discourses frame Big Data and AI as solutions to "otherwise intractable social, political and economic problems, and seem to promise efficiency, neutrality, and fairness" (p. 18). They argue that AI proponents often use spectacle to manufacture legitimacy, and emphasize what is "possible" rather than what is realistic. This framing obscures these fields' limits and tradeoffs, including the long-term outcomes and ripple effects that can result from the implementation of AI in the social world. Overdorf et al. (2018) share this concern about long-term outcomes and ripple effects. They describe how ML fairness literature's frequent presupposition of the algorithm and the environments in which it is deployed as fixed makes it "ill-equipped to prevent the negative effects that arise when agents evolve" (p. 3). One such example they provide is of predictive policing, wherein data is derived from communities that are disproportionately patrolled, creating a feedback loop that intensifies the surveillance of these communities. This point is similar to Selbst et al.'s (2019) understanding of the ripple effect trap that results from the abstraction of ML from the social contexts its design and implementation are embedded within (see section 3.3). Hegemonic approaches often fail to address the ways that the insertion of a given AI system into an existing social world will have transformative impacts on pre-existing behaviors and values.

## 4. Proposed Solutions

A variety of solutions have been proposed for addressing the limits and harms of hegemonic ML fairness research described above. However, not all solutions are equally well positioned to fundamentally disrupt the existing power dynamics and structural injustices historically and presently shaping the field. This is not to say that only highly disruptive solutions are useful, but it is to say that some solutions on their own are less likely to produce just outcomes for society's most marginalized. At their worst, however, some proposed paths forward can contribute to ethics washing or the legitimation of dominant systems of power more broadly. One such example is improving the efficiency and accuracy of algorithms for pre-trial risk assessment, which reinforces carceral logics and outcomes, and siphons off political will for more radical and necessary reforms, such as the ending of cash bail and pretrial detention (Green, 2019). This example is part of what Benjamin (2019) identifies as the New Jim Code: forms of technological practice that reproduce structural racism, and yet appear to offer greater impartiality. In what follows, the author describes a sample of both technical and non-technical solutions, and evaluates whether they are positioned to disrupt existing power dynamics and deeply entrenched forms in inequality in society. This is by no means an exhaustive list, but it is meant to provide examples for thinking through the possibilities and potential limits of different and adjacent approaches.

## 4.1  Causal Graphs

One proposed technical solution to the limits of existing AI fairness research is Mitchell et al.'s (2021) argument for incorporating causal graphs. These authors are working to find a place for quantitative modeling and quantitative measures for making ethical judgments amidst the critique that quantitative notions of fairness fail to address the structural conditions that sustain inequality. According to the authors, causal reasoning with clearly articulated goals could be used to design interventions and estimate the effects of decisions rather than to define fairness. Causal reasoning assesses cause-effect relationships between relevant attributes that lead to a decision (Hu, 2019). Mitchell et al.'s (2021) proposed technical solution recognizes that all aspects of designing AI tools are value-laden. Furthermore, the authors state, "it is crucial to involve members of the impacted community in the entire development process" (p. 158). Causal graphs on their own do not address how ML practitioners currently have a disproportionate amount of power in deliberations and debates concerning what society's objectives should be and how they should be achieved due to a range of social, institutional, and economic privileges (Burrell & Fourcade, 2021). It seems difficult to imagine a model-driven or automated decision making tool that coincides with democratic, deliberative, and just decision-making mechanisms unless explicitly designed with anti-oppressive goals and inclusive of the needs and perspectives of marginalized groups. This limitation also applies to Google's ML-fairness gym, which on the one hand provides tools for simulating system level dynamics, feedback loops, and other long-term effects of ML systems in order to better evaluate fairness considerations in relation to a given social context (D'Amour et al., 2020). On the other hand, however, it does not require that AI researchers work with community members or domain experts from other disciplines who might be better positioned to understand how agents' behaviors are institutionally conditioned and constrained by power dynamics within a given social context.

Additionally, Hu (2019) argues that recent attempts to overcome the limits of parity-based fairness metrics with causal diagrams are flawed, given the ways this method is often incompatible with an understanding of race as a social category. She writes, "a causal diagram that draws race as a singular node that independently *causes* downstream effects, or that sees race as something that can be isolated from confounders and mediators such as 'socioeconomic status' or 'high school performance' rests upon an incorrect theory of *what race is*—and why it matters in our society." She argues that counterfactual methods rely on a phenotype-only view of race. Key to her argument is that this framework allows for wrongful discrimination to be reduced to irrationality in decision-making rather than concrete social meanings and practices that emerge from a society organized around the production of racial stratification. Similarly, drawing from Hu's (2019) work, Malik (2020) argues that, "from a sociologically rigorous point of view, the idea that one could hold everything but somebody's race constant is incoherent: aspects of people are so deeply tied into how their racial identity is constructed, and the 'effects' of race so suffused into so many other variables, that a hypothetical 'race-switched' world is meaningless" (p. 16).

## 4.2 Intersectionality as Differential Fairness Classifier

Another technical solution that has emerged is to incorporate intersectionality into technical fairness measures. According to Foulds, Islam, Keya, and Pan (2018), an intersectional definition of AI fairness should satisfy the following criteria: "A) Multiple protected attributes should be considered; B) All of the intersecting values of the protected attributes, e.g. Black women, should be protected by the definition; C) We should still also ensure protection is provided on individual protected attribute values, e.g. women; D) The definition should protect minority groups, who are often particularly affected by discrimination in society; E) The definition should ensure the systematic differences between protected groups, assumed to be due to structural oppression, are rectified, rather than codified" (p. 3). To demonstrate the practicality of their differential fairness classifier method, they use data from COMPAS. While the paper nuances discussions of fairness with a greater consideration of how different systems of oppression intersect, the uncritical use of COMPAS to establish the efficacy of their methods speaks to the limitations of this approach. While it may be true that this intersectional approach to technical fairness has better outcomes for minority groups than subgroup fairness, making a tool for predicting the likelihood of recidivism intersectionally fair does not interrogate the problem formulation space's inherent bias towards incapacitation and its constitutive power dynamics.

## 4.3 Redirecting the Problem Formulation Space

Beyond new technical approaches to assessing fairness, other scholars have argued for the redirection of ML research such that the problem formulation space is explicitly geared towards the interrogation of systemic discrimination. Green and Hu (2018) cite Goel et al.'s (2016) use of ML for understanding racial disparities in New York City's (NYC) stop-and-frisk policy as an example of such work. In Goel et al. (2016), the authors deploy ML methods in order to intervene in the debate concerning stop-and-frisk policy's constitutionality and relationship to racial discrimination. They find that stop-and-frisk policies in NYC are both racially discriminatory and frequently do not meet the bar for reasonable suspicion. Based on these results, the authors argue that it is possible to design stop heuristics that would guide officers in only conducting stops that are statistically likely to result in the discovery of weapons (the most commonly occurring suspected crime in their dataset), and produce a more racially-balanced pool of stopped suspects. According to the authors, this effort "could temper public reaction to the policy, including resentment and distrust of police" (p. 383). While this article shows an example of ML research that directly grapples with questions of systemic discrimination and focuses its efforts on evaluating dominant institutions of power, it does not sufficiently engage with the larger sociotechnical context of US policing, including the criminalization of poverty and the structural conditions that produce criminogenic factors in the first place. Thus, the article makes assumptions about the likelihood that these police reforms will not only be adopted, but also un-problematically implemented. Ultimately, the function of such an approach could be to ethics wash policing itself, disarticulating it from its foundational relationship to racial capitalism in the US (Davis, 2003; Gilmore, 2007; Benjamin, 2019) and imperial power abroad (Shrader, 2019). As Moore (2019) and Abebe et al. (2020) caution, many AI for "social good" efforts are not sufficiently aimed at the root of problems, diverting attention from demands for structural change.

In the specific context of ML fairness debates concerning the US criminal legal system, Barabas et al. (2018) argue that instead of risk assessment tools grounded in regression and ML, which lend themselves to incapacitation rather than rehabilitation, a casual inference framework would allow for the consideration of criminogenic factors for more ethical intervention. These factors might include the existing criminal legal system itself and racialized and gendered covariates, as well as point towards ways to address the underlying social, economic, and psychological drivers of crime. This approach requires the design of "experimental conditions in which covariates are altered systematically to see if the alteration produces effect changes in the outcome variable" (p. 6), ultimately helping researchers identify what interventions in the criminal legal system work. They offer randomized control testing regarding psychological therapy in prisons as an example of what this might look like in practice, and invoke US institutional bodies like Institutional Review Boards (IRB) to address the counter-argument that there are ethical concerns with such an approach. Ultimately, the authors are shifting the conversation from the question of trade-offs regarding, for example, accuracy equity and predictive parity, towards deeper questions about "what the purpose of these tools is and should be in the first place" (Barabas et al., 2018, p. 7). They note that a combination of quantitative and qualitative methods, with greater researcher reflexivity in examining underlying assumptions, would help in ultimately understanding and responding to underlying drivers of crime. They are also not opposed to the use of ML for the development of diagnostic risk assessments, and think ML could be useful in identifying highly correlational covariates, which could then inform hypotheses on interventions.

While this approach takes into consideration the underlying structural conditions that contribute to criminalized behavior, it gives short thrift to the ethical concerns and power imbalances surrounding randomized control testing with prison populations (a group that is already significantly marginalized and without a range of protections). Several feminist and critical race studies researchers have critiqued the IRB as prioritizing institutional liability over justice considerations (see, for example, Cooky, Lindabary, & Corple, 2018; Sabati, 2019; Weinberg, 2020). It is also worth noting that this proposed solution might be an example of the solutionism trap (Selbst et al. 2019) in the sense that there have been decades of abolitionist work in a range of humanities and social science disciplines that have documented the social policies, legal reforms, and community resources needed to address the systemic poverty and racial inequality that feed the existing US criminal legal system, a system designed to make it harder, not easier, for people to access the support structures needed to live, work, and redress harm (e.g. Davis, 2003; Gilmore, 2007; Benjamin, 2019). Yet, more radical reforms have not received robust implementation as they pose fundamental challenges to the existing social order.

## 4.4  Fairness Checklists

Another proposed solution for remedying hegemonic approaches to ML fairness is the implementation of AI fairness checklists in order to empower individual advocates and formalize ad-hoc fairness practices. In the context of EU non-discrimination law, Wachter, Mittelstadt, and Russel (2021) propose a yes-no checklist for ML developers, deployers, and other users of automated decision-making systems to encourage these groups to justify fairness-related decisions and to incorporate technical fairness criteria that counter inequality. Madaio et al. (2020) also propose fairness checklists, but do not employ questions with "yes" or "no" answers, which they feel avoids the assumption that AI ethics can be addressed via purely technical solutions. They account for the limitations of checklists that have resulted in misuse, including in the airline industry and medical field, due to lacking organizational processes and cultures, practitioner motivation, and workflow alignment. In Madaio et al.'s (2020) co-design workshop with ML practitioners, participants expressed that checklists might help reduce anxiety or fear about social costs and encourage productive tensions and critical conversations. However, to be effective, participants felt checklists must be "aligned with existing workflows, supplemented with additional resources, and not framed as a simple compliance process" (p. 6). Some participants reported apprehension about the checklist approach as potentially producing bad practices, such as cutting corners or superficial completion and/or a reliance on overly procedural understandings of fairness.

While Madaio et al. (2020) bring up the risks of technosolutionism and ethics washing with checklist-based approaches, they are optimistic about the possibility of creating checklists that can mitigate the understanding of AI fairness as a purely technical consideration, if supported by the larger organizational culture and "designed to prompt discussion and reflection that might otherwise not take place" (p. 10). Wachter, Mittelstadt, and Russel (2021) express a similar weariness of technosolutionism, but an optimism concerning the capacities of checklists to prompt critical self-reflection. However, there are several possible limitations with checklist approaches to consider. These approaches often rely on "individual advocates" using checklists in order to push through AI fairness considerations, as opposed to ethics being embedded within, and the responsibility of, all levels of a given organization. Additionally, checklists, on their own, are unlikely to transform the limits of many existing legal frameworks and industry cultures, particularly those that are deeply entrenched in longstanding commitments to technological solutionism and the limiting of worker power. This is why scholars like Metcalf, Moss, and boyd (2019) are skeptical of checklists. Furthermore, there remains the risk of checklists being privileged over other efforts, such as regulation, precisely because they do not compel the redistribution of power in the design and implementation of AI tools.

## 4.5 Improving Data Collection Practices

In terms of the problems identified with dominant data collection practices in AI fairness research (see section 3.7), Jo and Gebru (2020) argue that ethical data collection practices for ML can draw from the institutional and procedural strategies and structures of archives and libraries for addressing "ethics, representation, power, transparency, and consent" (p. 10). Building on the practices of archives and libraries, the authors argue for researchers to have a public mission statement for guiding their data collection practice and encouraging contributions from other researchers. They also advocate for researchers to incorporate more open-ended input from communities through improved crowdsourcing and participatory collection measures. Furthermore, Jo and Gebru (2020) argue for the addition of multiple levels of review and record keeping for transparency and accountability, and to have a designated data collector role that is responsible for complying with ethics codes. While an interdisciplinary sub-field focused on "data gathering, sharing, annotation, ethics-monitoring, and record-keeping processes" (p. 2) alongside "governing bodies…to audit collectors" (p. 10) could have significantly positive impacts, the location of responsibility for ethics compliance within a "data collector" role, and relying on institutional/professional codes for ethical practice, have been critiqued as being inadequate for combating longstanding industry trends (Metcalf et al., 2019).

Elish and boyd (2017) also call for more critical, reflective approaches to data collection, and particularly, for AI researchers to account for how "their research practices might influence or distort the knowledge that results from their work" (p. 17). Elish and boyd (2017) also ask researchers to account for their own positionality as embodied and socially situated actors, which impacts the knowledge claims they make and how. Building on the methodological insights of anthropological ethnography, these authors feel that greater critical attention to the ways that cultural values and social and historical contingencies get embedded into data collection processes will result in more appropriate and effective design. Drawing from value-sensitive design, Dobbe et al. (2018) also argue for ML practitioners to take seriously how their social position shapes their epistemology and to engage in critical self-reflection in conversation with affected stakeholders. A more self-reflexive methodology for AI would be an impactful development, as it is of paramount importance that AI developers recognize that they are always already engaged in political and ethical judgment-making when building these tools. However, it is important to note that a more self-reflexive AI researcher methodology does not, on its own, upset power dynamics concerning who gets to fully engage in the problem formulation and research stages of AI development in the first place.

In order to promote greater transparency and accountability concerning ML dataset practices, Gebru et al. (2021) propose the creation of a standardized process for documenting the motivation, composition, funding, collection processes, distribution, and recommended uses of these datasets. They argue that these "datasheets for datasets" have the potential to "mitigate unwanted biases in machine learning systems, facilitate greater reproducibility of machine learning results, and help researchers and practitioners select more appropriate datasets for their chosen tasks" (Gebru et al., 2021, p. 2). While datasheets might help the ML community to be more reflective about their data practices, and potentially equip policy makers, consumer advocates, individuals included in datasets, and those who might be impacted by a given model with useful information for contesting unjust data uses, greater transparency does not always lead to accountability. As Ananny and Crawford (2016) argue, being able to "see" a system does not necessarily mean one is empowered to govern or change it. Datasheets, on their own, do not create mechanisms for addressing the power asymmetries that exist at all stages of AI's design, development, and deployment, nor necessarily curtail unethical data collection practices, but they could certainly be coupled with these efforts. Relatedly, Denton et al. (2020) have called for ML practitioners to reflexively interrogate how data pipelines are shaped by the decision making practices, motives, working conditions, methodologies, and embedded values of dataset developers, and have proposed a research program for promoting accountability and contestability in the development of benchmark datasets.

Hanna et al. (2020) also lay out possible avenues for addressing the harms that hegemonic approaches to data collection pose to marginalized groups, and draw from the insights of the social sciences in order to do so. To mitigate the ways that racial categories are taken up in AI fairness research such that oppressed groups are treated as interchangeable and the structural aspects of algorithmic unfairness get minimized (see section 3.4), the authors propose that algorithmic fairness researchers work to engage in context-specific understandings of race. Such an understanding would require considering the fact that race is unstable across time, place, and context. It would also require taking seriously the question of who is doing the classifying, for what purposes, and to what ends. Furthermore, they argue that algorithmic fairness research needs to recognize that measurements of race are reflective of the social properties that help produce, shape, and condition race. Thus, different dimensions of race, from racial self-classification to the race others believe you to be based on appearance or interactions, "might reveal different patterns of unfairness in sociotechnical systems" (p. 10). These categories and measurements should be critically assessed, and justification should be provided for the use of any racial schema used in research. Hanna et al.'s (2020) interdisciplinary approach is highly context-specific and process oriented, and it both centers and addresses the impacts of structural racism on classification systems. However, it is also possible that, like Elish and boyd's (2017) recommendation for more self-reflexive research practices, such an approach might be appropriated within the ML fairness community to function as a more elaborate discussion of the limits of a chosen classification system, without curtailing harmful ML research.

## 4.6  Protective Optimization Technologies

In response to hegemonic fairness frameworks that do not interrogate the fundamental goals of a given optimization system, as well as its negative externalities and long-term outcomes, Overdorf et al. (2018) propose the creation of "Protective Optimization Technologies" (POTs), inspired by creative user approaches to subvert problems with Waze routing, Uber wage suppression, and discriminatory targeted advertising. POTs "empower people and environments affected by optimization systems to intervene when OSPs [optimization system providers] fail to meet their needs" (p. 4). POTs are explicitly modeled on, and evaluate, the impact of optimization systems on populations and environments in order to broaden the scope of unfairness and influence system outcomes, such as by poisoning training data, crafting alternative inputs to counteract the optimization effect, and other adversarial machine learning techniques to reduce negative impacts on users and the environment. On their own, POTs can sometimes privilege individual modes of resistance over collective organizing. For instance, the University of Chicago SAND Lab's algorithm and software tool, *Fawkes,* helps individuals limit how unknown third parties track their images for improving facial recognition technology, as well as poison models that try to learn what that individual looks like (Shan et al., 2020). Furthermore, the development and implementation of some POTs, such as TensorClog (Shen, Zhu, & Ma, 2019)—a poisoning attack technique for increasing the privacy of user data from deep neural networks—require levels of technical knowledge that are not accessible to all. Additionally, while POTs can be used to help users avoid or disrupt harmful optimization systems, acts of modifying, subverting, or sabotaging an optimization system can also be used for antisocial ends.

### 4.7  Addressing Social Context

Regarding the issue of hegemonic ML fairness approaches failing to consider social context, Selbst et al. (2019) offer several ways of incorporating social context into ML fairness considerations, drawing from STS's sociotechnical framing and methodology. These authors propose: taking a process-oriented approach that involves "social actors, institutions, and interactions" (p. 60), including "local incentives and reward structures, institutional environments, decision-making cultures, and regulatory systems" (p. 64); modeling for these factors using normative values that are based on the social context; and being transparent in how these decisions get made and including those who will be most significantly impacted by these systems in determining appropriate norms. Selbst et al. (2019) also encourage technical researchers to acknowledge and account for the ways that existing power dynamics shape and condition how a technical system will operate and how its use will evolve in practice. This includes acknowledging and mitigating the ways that quantifiable metrics can be unconsciously privileged at the expense of other fairness considerations, and grappling with the ways that modeling/AI might not be the appropriate or best solution to a problem, particularly when "fairness definitions are politically contested or shifting" (p. 63) or when "the modeling required would be so complex as to be computationally intractable" (p. 63). Ultimately, these solutions require technical researchers to learn new skills or collaborate with social scientists, and to consider working with advocacy organizations or vulnerable populations directly. Selbst et al.'s (2019) solution is explicitly proposed to mitigate technological solutionism and exclusionary AI design practices, and this would hinge on the degree to which marginalized people are involved and empowered.

In response to calls for designers of algorithmic systems to take more process-oriented approaches that foreground algorithmic justice and include social context in fairness assessments, Barabas et al. (2020) draw from the anthropological practice of "studying up." Rather than directing research efforts towards examining disadvantaged subjects, studying up entails examining those who have agency, authority, and institutional power. Studying up also demands a consideration of how issues of power and domination impact the research process, including the knowledge claims that get produced, the research that gets funding and given access, and the positionality of researchers. In anthropology, the call to study up emerged in part due to the field's complicity in US imperialism in the 60s and 70s, such as through anthropologists' participation in Project Camelot, conducted at American University. The United States Army funded this counterinsurgency study in 1964 for the purposes of predicting and influencing social and political developments in Latin America in order to maintain US economic hegemony in the region. The project received international and national backlash, resulting in its cancellation (Solovey, 2001). Similarly, data science has faced several recent high profile controversies regarding its relationship to larger social and political struggles, including discriminatory outcomes in targeted advertising, automated image recognition, and AI chatbots (Barabas et al., 2020). Drawing from these recent controversies, Barabas et al. 2020 argue that "data scientists tend to uncritically inherit dominant modes of seeing and understanding the world when conceiving of their data science projects" (p. 170).

In order to illustrate what "studying up" looks like in practice in the context of ML fairness work, the authors evaluated how data science interventions concerning bail reform mirror the default epistemological and normative assumptions of the US criminal legal system. Instead, the authors argue that what to study, what to predict, and whose behavior to influence should be shaped in relation to evaluating "a judicial culture that permits excessive incarceration" (Barabas et al., 2020, p. 171). Their studying up involved collaborating with community organizers working on bail reform through conversations and a roundtable, which led them to focus on judges' agency and accountability in relation to state and federal laws protecting against excessive pretrial detention in the US. Ultimately, the researchers found a lack of needed data regarding defendant outcomes, which led them to refuse to try to reform predictive tools for bail on ethical grounds. Instead, they produced a judicial risk assessment tool that provides a "failure to adhere" score based on whether a judge is likely to adhere to the US Constitution regarding imposing unaffordable bail without due process. While their model surpasses the accuracy and validity of mainstream pretrial risk assessments, they say it is not intended for practical use, but rather, to help imagine alternative methods, discourses, and thus futures by "subjecting those in power to the very socio-technical processes which are typically reserved for only the poor and marginalized" (p. 174). Given the fact that community members helped guide the question framing, methodology, and outcomes of the development of the AI tool based on their experiences and needs, and given that the site of data collection does not render marginalized groups more vulnerable to repression because the target of collection is dominant institutions and social actors, Barabas et al.'s (2020) approach is positioned to redress entrenched power dynamics in ML fairness research.

## 4.8 Interdisciplinarity and Cross-Disciplinary Collaboration for New Methodologies

While some scholars have borrowed from the methodological insights of the social sciences, others have explicitly called for interdisciplinarity to inform a new methodology for the design of AI tools in order to produce just outcomes. Green (2019) calls for leveraging interdisciplinarity in order for algorithmic interventions to be evaluated against alternative reforms so that solutions for meaningful social change can be identified, and for designers to consider the insights of social thinkers and activists. Pasquale (2018) argues that engaging with interdisciplinary insights from social theory, critical race theory, and feminist theory would allow for "a more inclusive and critical conception of algorithmic accountability" in order to mitigate the continued entrenchment of power asymmetries under the pretenses of addressing bias. West et. al (2019) also argue that interdisciplinarity would allow for ML fairness research to go beyond technical debiasing and meaningfully address how AI is used in context. Similarly, Binns (2018) argues that algorithmic fairness considerations should account for how and why certain social circumstances exist through engagement with other disciplines, such as sociology, history, and economics. He argues that feature selection should be informed by an awareness of historical causes of inequalities and broader existing social structures. Considering these factors might also help with determining which disparities should take priority in fairness analysis and mitigation, such as focusing attention on the historic reasons for unequal base rates when evaluating the fairness of COMPAS. Relatedly, Malik (2020) imagines mixed methods research through collaborations with researchers with different toolsets and perspectives alongside greater researcher reflexivity, reciprocity with impacted communities, and humility, including holding space for the possibility that other methods might be most appropriate. Manjarres et al. (2021) have also called for a coordinated interdisciplinary effort, in particular to establish an "AI for Equity" field (AI4Eq) in response to the United Nations Agenda for Sustainable Development, drawing from ICT for Development.

Given the ways that interdisciplinary collaborations have the potential to challenge hegemonic computational epistemologies and practices, and particularly, problems of technological solutionism and abstraction, there is reason to believe that such collaborations would disrupt existing institutionalized power dynamics within ML fairness research. Yet, the question of which interdisciplinary frameworks are taken up for ML fairness considerations, and how they get leveraged, is paramount. For instance, Irani et al. (2010) have argued for the need to shift from developmentalist frameworks to STS and postcolonial computing frameworks that center "questions of power, authority, legitimacy, participation, and intelligibility." Mohamed et al. (2020) draw from Irani et al.'s (2010) critique when they envision decolonial AI rooted in solidarity rather than benevolence (see section 4.9). As Abebe et al. (2020) note, there is a longstanding and rich tradition of critical reflection in STS, Values-in-Design, and legal scholarship that some AI/ML practitioners have begun taking up over the past five years, but many of these interdisciplinary engagements have not been well aligned with broader efforts for social change. Examples of recent efforts to incorporate value sensitive design into ML include Umbrello and van de Poel's (2021) bringing together of value-sensitive design with AI for social good principles, arguing that there is a need for ML to be explicitly orientated towards socially desirable ends. They argue that these ends can be modeled on the United Nation's Sustainable Development Goals. However, it is not specified to what degree the participation of impacted communities figures into this approach, nor whether these approaches set out to substantively alter relations of power in society.

Abebe et al. (2020) provide a model for how computing can be an accomplice to efforts for structural change, even if insufficient on its own. They propose using computing to: measure social problems and diagnose how they manifest in technical systems; to generate public debate about how problems are being formalized for decision-making tools; to illuminate gaps, constraints, and flaws with existing datasets; to challenge entrenched policy thinking and harmful, inaccurate, and overstated technological capabilities through computational practice and advocacy work; and to use computing to call attention to long-standing social problems in ways that catalyze public engagement. Abebe et al. (2020) are attentive to the potential limits and drawbacks of these approaches, including the possibilities that these sorts of critical interventions will be appropriated or watered down by powerful stakeholders. Greene, Hoffman, and Stark (2019) share this concern about co-optation, demonstrating how several high-profile ethical AI vision statements engage with some of the criticisms of STS and other critical scholarly perspectives, but often limit ML fairness debates to appropriate design and implementation, not whether these systems should be built at all, or whether the "dominance of corporations on political processes with no democratic accountability" (p. 8) is a causal factor in harmful outcomes.

## 4.9  Participatory Design and Co-Development

In addition to interdisciplinarity, many scholars see robust participatory design measures as a method of significantly improving the development of AI tools in order to produce more just outcomes. Whittaker et al. (2019) for example imagine a shift in AI fairness discourse from bias and inclusion towards agency and control, where AI developers collaborate with disabled people who are empowered to define the terms of engagement and the order of priorities. This approach is seen as a way of challenging sociotechnical systems that exclude and oppress disabled people and other marginalized people. Participatory design processes that reimagine power relations between technologists and marginalized communities have the potential to significantly address what is most ethically concerning rather than what is most convenient to measure (Blodgett et al., 2020). Such an approach would make it possible for the design process to meaningfully leverage the knowledge that comes from people who occupy subjugated standpoints in society (Costanza-Chock, 2020). For Kalluri (2020), research should be oriented less around providing highly accurate information to decision makers and instead to serving the needs of marginalized communities, who are empowered to investigate, contest, influence, and dismantle AI. This approach would be AI "by and for the people" (Kalluri, 2020).

Drawing from decolonial theory, a body of work that actively seeks to disrupt colonial systems of power, Mohamed et al. (2020) propose AI co-development projects "driven by the agency, self-confidence, and self-ownership of the communities they work for" (p. 13). These projects would explicitly be designed for the purposes of exposing systematic biases and directly confronting colonizing knowledge systems and tools of expropriation. According to Mohamed et al. (2020), this work could be designed to support grassroots organizations that are already doing the work of generating intercultural dialogue, solidarity, and alternative modes of community, and by ensuring marginalized groups have meaningful decision-making powers over the AI tools' development. Such an approach would counter tendencies towards importing AI tools made in the West by Western technologists to non-Western contexts like Africa, which, as Birhane (2020) describes, can be both "irrelevant and harmful due to lack [of] transferability from one context to another," and also "an obstacle that hinders the development of local products" (p. 396). Such outcomes are in keeping with longstanding histories of imperialist oppression in contexts like Africa and India (Birhane, 2020; Sambasivan et al. 2021).

It is important to note that in order for participatory design to distinguish itself from participation-washing, those from marginalized groups who are subject to heightened levels of scrutiny and potential harm within many AI systems must have a significant degree of control over the development, deployment, and assessment of the AI tool. Sloane et al. (2020) argue that participation in ML can be carried out equitably by: recognizing participation as work that requires consent, opt-out options, and compensation; participation processes that are context specific and co-designed; and be genuine and long-term through transparency and knowledge sharing. This work necessarily requires de-centering Western notions of algorithmic fairness in non-Western contexts, including working with relevant culturally and historically contextualized categories, social structures, power dynamics, value-systems, and going beyond model fairness (Sambasivan et al., 2021). However, as Tucker and Yang (2012) caution, decolonization is frequently adopted as a metaphor divorced from the actual revolutionary project of restoring land and cultural, economic, and political freedom to colonized people. Adams (2021) argues that any effort to draw decolonial theory into AI must situate AI within the history of colonialism that informs the field's foundational epistemic assumptions regarding racial classification and notions of intelligence. The Indigenous Protocol and Artificial Intelligence Working Group is one such effort to imagine decolonial AI systems that prioritize Indigenous concerns and perspectives. For instance, the group's position paper offers guidelines for indigenous-centered AI that include Indigenous communities developing computational methods that reflect their own cultural practices and values, and ethically accounting for how hardware and software materials are extracted from and returned to the earth (Lewis, 2020).

## 4.10 Increased Democratic Deliberation and Regulation

In response to an absence of democratic deliberation in AI fairness considerations, McQuillan (2018) calls for more democratic forms of governance over the development and use of AI tools, and more specifically, for the creation of people's councils in contexts where people are significantly impacted by ML. He defines people's councils as "bottom-up, confederated structures that act as direct democratic assemblies" (McQuillan, 2018, p.7). Participants are empowered through inclusive, horizontal structures for collective action. McQuillan (2018) draws from the successes of people's councils in the US contexts of healthcare in the early 1990s and the Combined Shop Stewards Committee of 1976. The idea of people's councils reimagines who has the power to hold AI to account for unfairness and how by decentering technologists. However, as McQuillan (2018) himself notes, the efforts of people's councils are not always taken up by the institutions they seek to inform or challenge. This means that grassroots coalition-based movements, ideally bolstered by the support of AI fairness researchers, would be essential for pushing for the disruption of entrenched institutional power dynamics. McQuillian (2020) makes this point as well when describing the potency of tech worker dissent against facial recognition technology when carried out in conjunction with social movements.

Other scholars have explicitly called for regulation as a vehicle for disrupting existing power asymmetries that underpin hegemonic ML fairness approaches. For instance, Powles and Nissenbaum (2018) argue that what is needed is US policy attention to the "wholesale giveaway of societal data that undergirds A.I. system development" and to "disincentivize and devalue data hoarding with creative policies, including carefully defined bans, levies, mandated data sharing, and community benefit policies, all backed up by the brass knuckles of the law." Certainly, one can imagine the development of regulations that would help shift ML fairness from a set of voluntary principles and superficial technical adjustments to something that provides legally enforceable safeguards and restrictions against abuses of power. However, while regulation has the potential to disrupt existing power dynamics, regulatory frameworks can also be used to further consolidate and centralize power, assets, control, and/or rights in the hands of a minority, such as in the context of international AI governance standards and AI for international social development that often privilege Western interests (Mohamed et al., 2020; Sambasivan et al. 2021). Furthermore, regulations like the European Union General Data Protection Regulation (2016), a series of data protection rules implemented in 2018 for addressing data security and issues of consent, transparency, and accountability, have been criticized as being "too slow and too soft" (Bodoni, 2021). The more recent European AI Act (European Commission, 2021), a risk-based framework for regulating AI, has received similar challenges concerning its efficacy. Over one hundred civil society organizations have argued in a collective statement that the act is in need of amendments in order to ensure it is precise and robust enough to address structural imbalances of power ("An EU Artificial Intelligence Act," 2021). In addition to stricter regulation, Birhane (2021) argues that, "social pressure through organized movements, strong reward systems for technology that empowers the least privileged, and a completely new application of technologies" (p. 57) are best positioned to counter the ways that the AI/ML community's current incentive structures pressure even the most well-intentioned developers towards maintaining existing power dynamics.

## 4.11  Centering Justice

For some scholars, foregrounding theories of justice into AI fairness work would have a transformative impact on the degree to which ML fairness discourse engages with structural oppression and social justice considerations. For instance, Dencik et al. (2018) use the term "data justice" to highlight the significant relationship between data and social justice that goes beyond issues of data processing. Data justice takes as its starting point not the data system itself, but rather "the dynamics upon which data processes are contingent in terms of their development, implementation, use and impact" (Dencik et al., 2018). This process necessarily involves evaluating the socio-historical conditions, power structures, and relationships that produce injustice in the first place. It centers the experiences of marginalized groups, with the goal of alleviating unnecessary and unwanted suffering. Drawing from disability studies, Bennet and Keyes (2019) argue that questions of justice rather than questions of fairness should be centered in the context of disability and AI in order to resist reinforcing the existing powers of medical gatekeepers, and to prioritize the needs of multiply marginalized disabled people. Furthermore, Sambasivan et al. (2021) point out that there is a range of alterative moral foundations of cultural importance outside the context of hegemonic Western notions of fairness, including restorative justice, which could provide more potent frameworks for redressing algorithmic harms. Restorative justice emphasizes repairing harm through community-based processes that involve those who have been harmed and those who take responsibility for the harm working together to determine solutions that promote accountability and collective healing. Benthall (2018) and Green (2018b) have both called for the explicit inclusion of justice at the ACM Conference on Fairness, Accountability, and Transparency as a core component of its research agenda. However, Gangadharan and Niklas (2019) argue that most fairness, accountability, transparency, and data justice studies approaches are techno-centric and process-oriented, when what is needed is an evaluative ethical framework that decenters technology. According to the authors, such an approach would "position sociotechnical systems of discrimination alongside other modalities and offer nuance into what injustice discrimination causes, who is affected, and how discrimination can be remedied" (Gangadharan & Niklas, 2019, p. 855). While Gangadharan & Niklas (2019) raise an essential point about the constraints of techno-centrism, it remains possible to imagine design processes for AI tools that are explicitly geared towards anti-oppressive ends. Such ML research might support both technical and non-technical solutions, and empower groups working to build a more just and equitable world. The author will return to this point in Section 5.

## 4.12 Diversifying the AI Field

Finally, any overview of solutions to the limitations of hegemonic ML fairness approaches must address the AI field's fundamental lack of robust racial and gender diversity. Both Guillory (2020) and West et al. (2019) argue that the AI sector's lack of diversity is deeply linked to the prevalence of biased and discriminatory AI tools. Guillory (2020) points out that racial bias pervades the AI field due to racial bias in who gets admitted to the field, who gets mentored, who is turned to for collaboration, what research topics get emphasized and de-emphasized, who gets hired in academia, and who is given adequate financial and community support. Guillory (2020) also argues that the field's complicity in the harms that repressive technological systems produce within Black communities can further drive marginalized people from the AI field. In response to the limits of both conventional workplace diversity and tech debiasing initiatives for addressing racial and gender discrimination, West et al. (2019) argue that there needs to be a deeper engagement with "workplace cultures, power asymmetries, harassment, exclusionary hiring practices, unfair compensation, and tokenization" (p. 3). Dominant "pipeline studies" often study primarily white women's individual psychology as a barrier to their full participation in computer science, as opposed to studying the barriers imposed by institutions, including their cultures and unfair distributions of power and privilege (West et al., 2019). However, as the authors and Hampton (2021) note, diversifying the tech workforce will not necessarily address the deeper structural challenges that AI poses to communities, which is why a power-centered framework is necessary for examining the relationship between technological development and the lived experiences of marginalized people.

Such a power-centered framework would require that the ML community critically attend to algorithmic tools' reproduction of longstanding histories of scientific racism and use of harmful systems of classification (Coalition for Critical Technology, 2020). Furthermore, this framework would need to account for the deliberate, historical and present-day pushing out of women from computing, with disparate consequences for Black women in particular (e.g., Hicks, 2017; Noble & Roberts, 2019; Birhane & Guest, 2020; Turner, Wood, & D'Ignazio, 2021). Central to such an approach would also be questions concerning who is harmed, who benefits, and who gets to decide in a given ML application context, grounded in an analysis that prioritizes justice considerations. This analysis would need to include but go beyond a politics of representation within the AI field and interrogate the ways that the field's very formation and dominant epistemic frameworks are deeply embedded with logics that sustain social and economic inequality, and often prioritize industry and military imperatives rather than community needs.

## 5. Discussion

This survey has argued that hegemonic computational approaches to ML fairness research currently perpetuate and amplify many of the existing unequal power dynamics in society. At present, most ML tools serve to optimize a given system, rather than shift the politics within those systems towards social justice (Green, 2018b). At minimum, this review conveys the need for modeling social actors, institutions, and their interactions in a given social context so that the ML tool's function across a social system is rendered observable, and for using fairness criteria that are context-specific and grounded in the voices, experiences, and needs of the marginalized people. The redesigning of AI tools with the full participation of people who are intersectionally disadvantaged under capitalism, white supremacy, heteropatriarchy, and colonialism, and the meaningful incorporation of intersectional feminist thinking into the design and evaluation of technical standards and bias audits, could improve AI outcomes for marginalized groups in contexts where data gaps for minoritized people are systematically contributing to harm (Costanza-Chock, 2018; Adamson & Smith, 2018). However, closing representational data gaps will be insufficient on its own to overcome the unequal social and economic conditions that produce these gaps in the first place, and that condition how "fixing" them will actually impact the lives of marginalized people in practice.

The ML fairness community must also be weary of not prioritizing quantifiable fairness metrics at the expense of other fairness considerations for the sake of building a workable AI tool, and this should include the consideration of whether a given AI tool should be built at all in a world deeply conditioned by entrenched power asymmetries. As D'Ignazio and Klein (2020) argue, tracing linkages between digital technologies and "historical and ongoing forces of oppression can help us answer the ethical question, should this system exist?" (p. 13). Whether to build an AI tool concerns a range of ethical and social issues that also require inclusive and critical discussion, and in cases of grave harm, some researchers have actively intervened. As just one example, in the wake of the police murder of George Floyd in Minneapolis, Minnesota, mathematicians at several US universities called for a boycott on collaborations with police departments for developing predictive policing software, arguing that these collaborations create a "'scientific' veneer for racism" (Linder, 2020). Their open letter, submitted to the *Notices of the American Mathematical Society* in June 2020, has since been signed by over 1,500 researchers from a range of universities within and outside of the US context (Linder, 2020).

While not all of the reforms proposed in Section 4 are "operationalizeable," this does not mean that these approaches are outside the frame of what the AI/ML fairness community can do or call for. To provide an example, a research team can set out to build an AI tool using as many sociotechnical considerations as possible in a world rife with injustice. Such an understanding necessarily requires interdisciplinarity and forming peer relationships with relevant communities at all stages, from research, to design, to implementation, to evaluation. This can be difficult or resource-intensive work, and it requires negotiating different epistemic frameworks, as well as the active mitigation of the unequal power dynamics that surface in collaborative work between participants who hold different degrees of institutional and social privilege. As Abdurahman (2019) explains, peer relationships are predicated on algorithmic designers critically addressing "whose language is used, whose viewpoint and values are privileged, whose agency is extended, and who has the right to frame a 'problem.'" Assessing the tool will require having a critical understanding of how existing actors operate in the given social context, which means having to grapple with power, privilege, and structural inequality. Ultimately, a model may technically produce desired outcomes according to the research team's value-laden criteria, but given what they and their community interlocutors know about the social context, it may be unlikely to be applied in a fair way long-term unless there is also a formal and robust process to contest these outcomes that impacted people are able to access and meaningfully leverage. Calls for—and participation in—the development of complementary social, political, and legal processes to ensure just outcomes must be treated as necessary to the work of ML fairness.

Perhaps, then, and as many of the scholars in this survey article have argued, it is time for a wholesale rethinking of what it means to do ML fairness work. For instance, whom does it serve to call an AI tool built on exploitative labor conditions and predatorily inclusive data collection practices "fair"? What might it look like to reorient ML fairness research around the goals of evaluating institutions' production of structural advantages, and of eliminating relations of domination? Whose voices, experiences, and epistemic frameworks are included in the design process, and in what ways do structural inequalities and historical patterns of oppression shape a given ML tool's distribution of risks and rewards? What might it look like for ML fairness research to engage in a deliberate, self-conscious reckoning with computer science's historical and present day complicity in injustice, including through disciplinary knowledge gaps, paternalism, and perpetuations of harmful pseudoscience (Elish & boyd, 2017; Brihane & Guest, 2020)? How might the AI field transform if this reckoning were to form the very foundation of computer science education? What would it mean to judge the "goodness" of an AI tool if "good" were defined as that which gives people the ability to "exist, survive, thrive, and change in positive ways of their own choosing" (Malik, 2020, p. 30)? How might ML fairness research be developed in solidarity with community and industry movements for progressive technological reform, for the rights of tech workers, and for greater agency, self-determination, and democratic control for marginalized people over AI tools?

There are existing models available for robust participatory research that center questions of power, justice, and community needs in the context of digital systems (e.g., "Our Data Bodies," n.d; "Design Beku," n.d.), and it is promising to see the rise of interdisciplinary workshops explicitly geared towards justice-oriented AI research at major computing conferences where productive tensions, collaborations, and alternative approaches might emerge (e.g., "Resistance AI Workshop," 2020). Yet, as this review has argued, a great deal of existing ML fairness research continues to entrench dominant relations of power and subordination, deploying ideas of participation, inclusion, and fairness to encourage the diffusion of algorithmic tools that optimize an unjust social order. Anything less than an anti-oppressive approach to AI is poorly equipped to support the societal changes necessary for a more equitable society. An anti-oppressive approach to AI would involve explicit commitments to undoing the unequal distribution of social, economic, and political power shaping the AI field and contributing to the systematic advantaging of dominant groups at all stages of the AI lifecycle. This requires not only bridging divides between different epistemic communities, but also aligning ML fairness work with existing, historically longstanding, and international struggles for just institutions and community relations.

## Acknowledgements

The author is grateful to Yan Zhou, Rakin Haider, Tom Doyle, Chris Clifton, Christopher Yeomans, Murat Kantarcioglu, and Blase Ur for their helpful feedback and recommendations during the drafting of this article, and to the reviewers and editorial team at *JAIR* for their generative comments during the peer review process. This article is based upon work supported by the NSF Program on Fairness in AI in Collaboration with Amazon under Award number 1939728, "FAI: Identifying, Measuring, and Mitigating Fairness Issues in AI."